**Title:** Underutilization of Syntactic Processing by Chinese Learners of English in Comprehending English Sentences, Evidenced from Adapted Garden-Path Ambiguity Experiment

Jiapeng Xu

**Abstract:** Many studies have revealed that sentence comprehension relies more on semantic processing than on syntactic processing. However, previous studies have predominantly emphasized the preference for semantic processing, focusing on the semantic perspective. In contrast, this current study highlights the underutilization of syntactic processing, from a syntactic perspective. Based on the traditional garden-path experiment, which involves locally ambiguous but globally unambiguous sentences, this study's empirical experiment innovatively crafted an adapted version featuring semantically ambiguous but syntactically unambiguous sentences to meet its specific research objective. This experiment, involving 140 subjects, demonstrates through descriptive and inferential statistical analyses using SPSS, GraphPad Prism, and Cursor that Chinese learners of English tend to underutilize syntactic processing when comprehending English sentences. The study identifies two types of parsing underutilization: partial and complete. Further exploration reveals that trial and error in syntactic processing contributes to both. Consequently, this study lays a foundation for the development of a novel parsing method designed to fully integrate syntactic processing into sentence comprehension, thereby enhancing the level of English sentence comprehension for Chinese learners of English.

# 1 Introduction

The conventional focus in the study of semantic and syntactic processing in sentence comprehension is their temporal sequence (Wei et al. 2024). Some argue that syntactic processing precedes semantic processing (Deniz 2022; Veldre and Andrews 2018), while others believe the two processes are intertwined (Yang et al. 2021; Zhu et al. 2018). However, an intriguing circular logic observed among Chinese learners of English shifts our interest from the relationship between semantic and syntactic processing (see Figure 1a) to the timing of syntactic processing in relation to sentence comprehension (see Figure 1b). Wang (2020) has observed that Chinese learners of English often attempt to comprehend sentences before parsing them. This discrepancy between common practice and widely accepted belief reflects a circular logic: syntactic processing used to aid sentence comprehension has to rely on prior sentence comprehension. After conducting rigorous experiments with statistical analysis, Author (2024) confirmed that syntactic processing frequently follows sentence comprehension in Chinese learners of English (see Figure 1c).

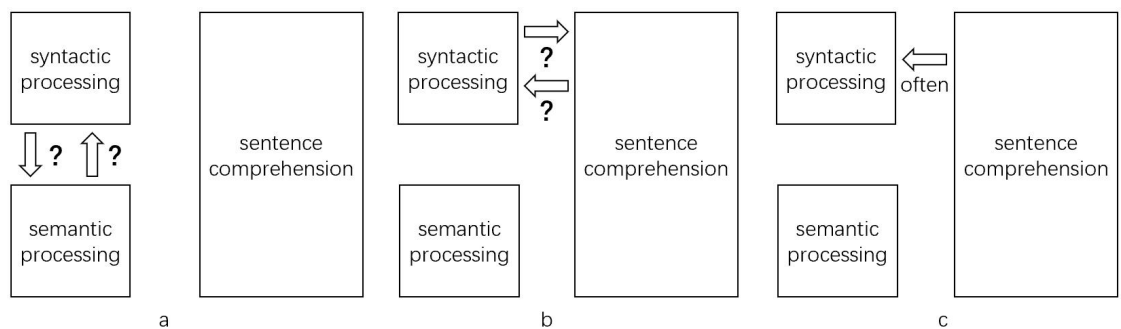

Figure 1 a. Time course of semantic and syntactic processing; b. Time course between syntactic

processing and sentence comprehension; c. Syntactic processing often follows sentence comprehension

Author's findings have led to the current study. It is widely believed that syntactic processing aids sentence comprehension (Choi and Zhang 2021; van der Burght et al. 2021). This suggests that parsing should be employed before and for sentence understanding. However, parsing methods based on prior sentence comprehension, as revealed by Author, is ineffective for understanding sentences. This suggests that grammar teaching plays a controversial role in foreign language education and may therefore be a source of confusion for student teachers (Graus and Coppen 2016). This raises the question of whether Chinese learners of English fully utilize syntactic processing for sentence comprehension. It has been demonstrated from various perspectives that we tend to rely more on semantic processing than syntactic processing during sentence comprehension. Cognitive neuroscience reveals that within the language system, lexico-semantic information is represented more robustly compared to syntactic information (Fedorenko et al. 2020; J. Wang et al. 2020). In electrocorticography, researchers have noted consistently stronger responses in language responsive electrodes when presented with conditions solely containing lexico-semantic information compared to conditions containing only syntactic information (Fedorenko et al. 2016). Pragmatically, this inclination towards lexico-semantic processing aligns with the perspective that the primary objective of language is communication, namely, the conveyance of meanings between individuals (Gibson et al. 2019; Hahn et al. 2020; Kirby et al. 2015). Syntactically, it aligns with the observation that the bulk of our linguistic knowledge revolves around lexical semantics, focusing on word meanings, while only a small amount is required to retain our syntactic knowledge (Mollica and Piantadosi 2019). While many studies reveal our utilization of semantic and syntactic processing in sentence comprehension, they typically emphasize a preference for semantic processing. Few studies, however, examine this issue from a syntactic perspective, which sheds light on the underutilization of syntax, its functions, and underlying mechanisms.

A specific illustration of the underutilization of syntactic processing in sentence comprehension is found in sentences containing local garden-path ambiguities (Christianson et al. 2001; Qian et al. 2018). For instance, Christianson et al. (2001) explored how individuals comprehend a locally ambiguous but globally unambiguous sentence "While Anna dressed the baby that was cute and cuddly played in the crib". The majority of participants believed that Anna dressed the baby. However, this interpretation contradicts the global sentence comprehension. This suggests that we often fail to fully utilize syntactic processing during comprehension. If the participants had thoroughly analyzed the sentence's grammatical structure—"While Anna dressed, the baby that was cute and cuddly played in the crib"—before deciding whether Anna dressed the baby, they would have realized she did not. Otherwise, the remaining clause, "that was cute and cuddly played in the crib", would lack coherence and fail to convey a coherent grammatical structure. However, the evidence from these locally ambiguous but globally unambiguous garden-path sentences is not entirely convincing as confirmation of the underutilization of syntactic processing, since the logic applies not only at the syntactic level but also at the semantic level. Specifically, we can also infer from the aforementioned experiment that we often fail to fully utilize semantic processing during comprehension. If the participants had thoroughly analyzed the sentence's semantic meaning—"While Anna dressed, the baby that was cute and cuddly played in the crib"—before deciding whether Anna dressed the baby, they would have realized she did not. Otherwise, the remaining clause, "that was cute and cuddly played in the crib", would lack coherence and fail to convey a coherent sentence meaning.

Moreover, previous theories do not assert that syntactic processing is underutilized in sentence comprehension among Chinese learners of English. However, addressing the underutilization of syntactic processing in sentence comprehension is crucial for Chinese learners of English due to their unique characteristics. Language specificity refers to the unique characteristics of each language, including syntactic differences between the source language (SL) and the target language (TL) (Gile 1997). A key sign of language specificity is the difference in word order between SL and TL (Gile 2005 2009; Li 2015). Research on different language pairs has shown empirical evidence that word order asymmetry adversely affects interpreting performance (Gile 2011; Seeber et al. 2012), including in English-Chinese translation (Ma et al. 2021; Ma and Li 2021). This presents challenges for Chinese learners of English in understanding English sentences compared to English L1 speakers (see Figure 2). Syntactic processing is considered essential for determining word order (Casado et al. 2005; Weyerts et al. 2002), highlighting the significance of syntactic processing for Chinese learners of English in understanding English sentences. Therefore, this current study focuses on Chinese learners of English, with the research question investigating whether syntactic processing is fully utilized in English sentence comprehension.

| Source language: His function is analogous to that of a judge, (who) must accept the obligation. |
|---|
|                 1     2     3       4      5  6  7 8  9        10    11    12     13 |
| Target language: 他的 功能 与 一位 必须 接受 这项 义务 （的） 法官 的 功能 是 相似的。 |
|                  1    2   5    8    10  11  12   13           9  7  6   4 |

Figure 2 The word order in comprehending an English sentence by English speakers corresponds to the transformed word order in comprehending it by Chinese speakers. The words "who" and "的" in the brackets have no transformed counterparts as they merely play a syntactic role in the sentences

    Furthermore, no studies have thoroughly investigated how syntactic processing is underutilized in sentence comprehension, and few have identified the reasons behind this underutilization. While some studies describe the influence of reading comprehension on syntactic analysis (Lee et al. 2015), they do not depict how the underutilization of syntactic processing arises due to parsing's reliance on sentence comprehension. While some studies attribute the underutilization of parsing to the limited effectiveness of existing methods (Fedorenko et al. 2020), given the inherent connection between syntax and meaning (Fedor et al. 2012), they do not explain why and how syntactic processing is linked to sentence meaning. Addressing all these gaps is the key contribution of this study. It not only advances linguistic theory on the time course of syntactic processing and sentence comprehension but also has practical

significance by laying the foundation for integrating parsing into sentence comprehension for Chinese learners of English.

The novelty of this study lies in its perspective of syntax's weak role in sentence comprehension and the focus on second language processing. In contrast, previous studies have primarily focused on first language processing, highlighting a preference for semantics. This shift is motivated by two factors. Theoretically, syntactic processing tends to lag behind sentence comprehension among Chinese learners of English (Author 2024), implying that current parsing methods are not effectively aiding comprehension and thus are underutilized. Practically, cognitive synergy entails the collaboration of various cognitive processes within a single system, aiding one another in overcoming internal processing barriers (Everitt et al. 2017). Syntactic processing, as one of the two primary cognitive processes in sentence comprehension, when underutilized, leads to a reduction in cognitive synergy. Therefore, it is significant to fully integrate parsing into English sentence understanding for Chinese learners of English. Another novel aspect of this study is its innovative use of a widely adopted experimental paradigm: the garden-path ambiguity. The traditional paradigm of garden-path ambiguity has the potential to provide evidence for the underutilization of syntactic processing. However, it reveals a mixed underutilization in both syntactic and semantic processing, as illustrated above. To reach a conclusion focused exclusively on syntactic underutilization, this study crafted an adapted set of garden-path sentences, shifting from the original locally ambiguous but globally unambiguous sentences to a revised version featuring semantic ambiguity but syntactic unambiguity. This adaptation was informed by prior analyses of sentences exhibiting syntactic ambiguities (Togato et al. 2017).

The objective of this study is to initially confirm the underutilization of syntactic processing in English sentence comprehension among Chinese learners of English. Subsequently, it delves deeper into the specific ways Chinese learners of English parse and the logic strategies through which they underutilize syntactic processing in English sentence comprehension. These efforts have both theoretical and practical implications. Theoretically, this study strengthens and expands on the argument that syntactic processing often follows sentence comprehension among Chinese learners of English by demonstrating their underutilization of parsing in English sentence comprehension. Practically, this current study allows us to develop novel cognitive parsing methods in the future to help them fully harness syntactic processing in English sentence comprehension for enhanced cognitive synergy.

## 2 Methods

### 2.1 Participants

The experiment involved 140 Chinese learners of English. Participants were randomly selected from a diverse pool representing 67 different majors across 12 universities, spanning disciplines such as mathematics, physics, chemistry, medicine, economics, management, law, politics, sociology, and more. Of the 140 valid samples, 110 were female and 30 were male. The selection criteria for participants were as follows: Firstly, they must be Chinese individuals learning English as a second language, aligning with the study's research goal. Secondly, they must be university students, indicating they have received comprehensive instruction in English grammar and are familiar with syntactic rules for reading comprehension of common materials.

Additionally, participants must be enrolled in non-language-related majors to ensure the experiment's results were generalizable and externally valid. Moreover, participants must have no direct involvement in the research beyond their role as study subjects, preserving the objectivity of the experiment's results. These measures were implemented to uphold the validity and reliability of the empirical experiment.

The 140 valid participants exhibit proficiency in English. All participants have undergone the College English Test (CET), a standardized English proficiency examination for non-English major students in China. The mean CET score among the 140 subjects was approximately 552 out of 710, with a standard deviation of around 55. Scores ranged from a minimum of 399 to a maximum of 678 (see Figure 3a). Given the moderate sample size, the Anderson-Darling test of normality (Anderson and Darling 1952) was employed to assess the normal distribution of participants' CET performance. The p-value from the Anderson-Darling test was approximately 0.3814 (>0.05), indicating that the data were normally distributed (see Figure 3b). This suggests that these participants are suitable for the experiment in terms of language proficiency. The detailed data pertaining to the gender, university, grade, major, and CET score of each valid participant are available in the Appendix File located at https://figshare.com/articles/dataset/Appendix_File/26023042.

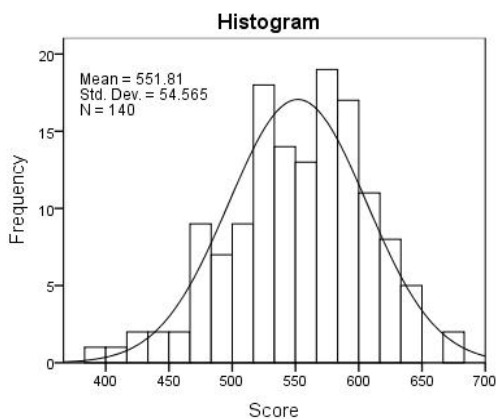
a

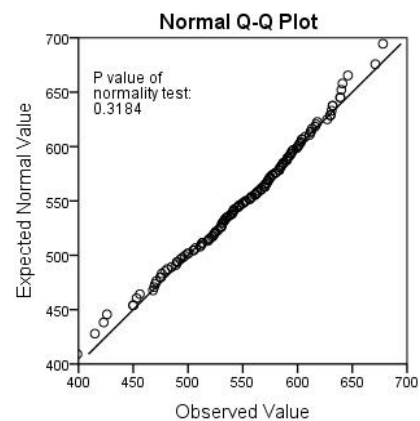
b

Figure 3 a. Histogram of the 140 participants' performance in College English Test (CET); b. Normal Q-Q plot of the 140 participants' performance in College English Test (CET) and the result of Anderson-Darling test of normality

*2.2 Hypothesis*

Given the analysis in the Introduction, this current study proposes the hypothesis that Chinese learners of English underutilize syntactic processing for comprehending English sentences.

*2.3 Design*

Although empirical observations from cognitive neuroscience studies suggest that responses to syntax are weaker than those to semantics (Fedorenko et al. 2020; J. Wang et al. 2020), we argue that these neuroscientific experiments provide weak evidence for the underutilization of syntactic processing, in that the neurocognitive response of the human brain to syntactic processing might be inherently weaker than its response to semantic processing, rather than being caused by the brain's underutilization of syntactic analysis. Therefore, the current research resorts to an experiment, relying on subjects' thought responses rather than neural responses, to confirm the possibility of the underutilization.

Inspired by the widely adopted design of garden-path ambiguities in numerous experimental studies e.g., (Christianson et al. 2024; Fujita 2021; Karsenti and Meltzer-Asscher 2022; Paape and Vasishth 2022; Pontikas et al. 2024), where sentences are locally ambiguous but globally unambiguous, we recognized that these experiments can offer evidence of underutilization in syntactic processing during comprehension. However, such evidence is limited as it does not specifically target syntactic underutilization. Rather, it points to the mixed underutilization of both semantic and syntactic processing, as elucidated in the Introduction. To address this weakness, we have modified the existing design by developing a new type of sentences that transition from being locally ambiguous but globally unambiguous to being semantically ambiguous but syntactically unambiguous. This approach allows us to specifically investigate the focus of this study: whether syntactic processing is underutilized during sentence comprehension.

The modified design consists of six paired sentences. All paired sentences share a common, simple syntactic structure, which is syntactically unambiguous, to help subjects comprehend them correctly and readily. However, one sentence within each pair is semantically ambiguous, allowing for more than one interpretation, while the other is semantically unambiguous, allowing only one interpretation. Consequently, if participants employ syntactic processing in sentence comprehension, their performance should exhibit insignificant differences between the paired sentences, indicating the likelihood of comprehending both. Conversely, if they predominantly rely on semantic processing, underutilizing syntactic processing for sentence interpretation, their performance would significantly differ between the paired sentences, with a tendency to comprehend the semantically unambiguous one but struggle with the other.

Two variables were involved in designing the experiment: Context (the independent variable) and Comprehension (the dependent variable). In each sentence pair, the two sentences share a common syntactic structure. For example, "The company plans to expand it <u>company-wide</u>" versus "The employee is known <u>company-wide</u>". However, the shared syntactic structure is presented within different contexts (the independent variable). The first context, "<u>The company plans to expand it</u> company-wide", renders two possible comprehensions (the dependent variable) to the sentence, making it semantically ambiguous if one relies solely on semantics and ignores syntactic cues: (1) The company plans to expand its scope, or (2) the company plans to expand something within the company. In contrast, the second context (the independent variable), "<u>The employee is known</u> company-wide", renders only one comprehension (the dependent variable) to the sentence even if understood semantically without syntactic consideration: Everyone in the company knows the employee. In conclusion, the inclusion of only two variables suggests that this experimental design has successfully controlled for the influence of confounding variables.

To draw robust conclusions, the experiment was approached from two different perspectives: correctness and confidence. Typically, higher correctness in responses was associated with superior performance. Similarly, increased confidence in answering questions correlated with better performance.

## *2.4 Materials*

To avoid bias, these six pairs of sentences were examined using various question formats, including multiple-choice questions, translation tasks, short-answer questions, and true/false questions.

Question 1: Interpret the meaning of the sentence "Every cat with an owner will run a small-scale study".

A. The men will run a study.

B. The cats will run a study.

C. Both the men and the cats will run a study.

D. Neither the men nor the cats will run a study.

Question 2: Interpret the meaning of the sentence "The woman with a glass looked at him".

A. The glass looked at him.

B. The woman and the glass looked at him.

C. The woman looked at him.

D. Neither the woman nor the glass looked at him.

Question 3: Translate the following sentence from English into Chinese "The company plans to expand it company-wide".

Question 4: Translate the following sentence from English into Chinese "The employee is known company-wide".

Question 5: According to the following sentence, what increases by nearly 20%? "The probability that the company has to restate earnings increases by nearly 20%".

Question 6: According to the following sentence, what increases by 2 degrees Celsius? "The temperature that is associated with heart rate decreases by 2 degrees Celsius".

Question 7: According to the following sentence, is it correct that assisting the homeless population is difficult? "Finding ways to assist the homeless population is difficult".

Question 8: According to the following sentence, is it correct that buying food is illegal? "Stealing money to buy food is illegal".

Question 9: Translate the following sentence from English into Chinese "Economists are warning of rising inequality and the increasing power of inherited wealth".

Question 10: Translate the following sentence from English into Chinese "The inherited wealth is about 100 million dollars".

Question 11: According to the following sentence, are there price managers? "These rules say they must value some assets at the price a third party would pay, not the price managers and regulators would like them to fetch".

Question 12: According to the following sentence, are there movie Jashon? "They are discussing the book introduced by Professor Michael, not the movie Jashon and his brother have watched recently".

The sentences involved in these questions are either selected from magazines such as *Business Week* or are specially crafted. The materials are provided in the Materials and Stimuli File, accessible at https://figshare.com/articles/dataset/Materials_and_Stimuli_File/27636918?file=5031 4413.

The 12 questions are organized into 6 pairs to form matched items for a comparative experiment.

In Pair 1 (Question 1 and Question 2), the participants were tasked with answering two multiple-choice questions based on a pair of sentences. Both sentences contained a prepositional phrase involving "with" after the subject of the sentences so that they were comparable. "Every cat with an owner" in Question 1 and "The woman with a glass" in Question 2 serve as the main stimuli. When comprehending "Every cat with an owner will run a small-scale study" (SEM AMB $_+$ SYN AMB $_-$) purely semantically, interpretations for Chinese learners of English could include the cats running a study, the men running a study, or both the cats and men running a study.

However, a syntactic analysis allowed them to explicitly determine that the sentence implied "the cats will run a study", as prepositional constructions such as "with an owner" cannot serve as the subject. In contrast, "The woman with a glass looked at him" (SEM AMB -  SYN AMB -) is unlikely to be misunderstood even if syntactic processing is not involved because it is semantically odd to believe that a glass looked at a person.

In Pair 2 (Question 3 and Question 4), the participants were tasked with translating a pair of sentences from English to Chinese. To ensure the comparability of the two sentences, both sentences contained the keyword "company-wide", which serves as the main stimulus. If participants do not leverage syntactic processing for sentence comprehension, "The company plans to expand it company-wide" (SEM AMB + SYN AMB -) is likely to be either inaccurately translated into Version 1: "这家公司想要扩大公司范围" (The company wants to expand the company itself) or correctly translated into Version 2: "这家公司想要在公司内部推广它" (The company wants to expand it within the company). However, if participants employ syntactic processing for sentence interpretation, the sentences is likely to be correctly translated, because company-wide is an adverb which is used to modify the verb of the sentence. In contrast, "The employee is known company-wide" (SEM AMB - SYN AMB -) is likely to be accurately translated only as "这家公司的所有成员都知道这个员工" (All the members of the company know the employee), as only this version makes semantic sense.

In Pair 3 (Question 5 and Question 6), participants were tasked with answering two questions based on a pair of sentences. The two sentences shared a common syntactic structure composed of "NP + VP + AP", allowing for meaningful comparison. "Earnings increases by nearly 20%" in Question 5 and "heart rate decreases by 2 degrees Celsius" in Question 6 serve as the main stimuli. The sentence "The probability that the company has to restate earnings increases by nearly 20%" (SEM AMB $_+$ SYN AMB $_-$) may be correctly understood but is also susceptible to being misunderstood as indicating a nearly 20% increase in earnings if syntactic processing is not fully employed. In contrast, "The temperature that is associated with heart rate decreases by 2 degrees Celsius" (SEM AMB $_-$ SYN AMB $_-$) is unlikely to be misconstrued as implying a decrease of 2 degrees Celsius in heart rate, even if syntactic processing is abandoned for sentence interpretation, because it sounds odd semantically.

In Pair 4 (Question 7 and Question 8), participants were assigned the task of making a true or false judgment based on a pair of sentences. Both sentences shared the same syntactic structure, which was conducive to meaningful comparison. "Finding ways to assist the homeless population is difficult" in Question 7 and "Stealing money to buy food is illegal" in Question 8 serve as the main stimuli. The sentence "Finding ways to assist the homeless population is difficult" (SEM AMB $_+$ SYN AMB $_-$) is prone to being misunderstood as indicating that assisting the homeless population is difficult because it sounds semantically sound, aside from being correctly understood as finding ways being difficult, if syntactic processing is not employed, probably leading to the oversight of "finding ways" as the actual subject of the sentence. In contrast, "Stealing money to buy food is illegal" (SEM AMB $_-$ SYN AMB $_-$) is unlikely to be misconstrued as implying that buying food is illegal even if syntactic processing is not involved in sentence interpretation, because such an interpretation would be semantically weird.

In Pair 5 (Question 9 and Question 10), participants were tasked with translating a pair of sentences from English to Chinese. To ensure comparability, both sentences shared the same noun phrase "inherited wealth", which serves as the main stimulus. If subjects do not employ syntactic processing for sentence comprehension, such as "...the increasing power of inherited wealth" (SEM AMB $_+$ SYN AMB $_-$), it is likely to be either inaccurately translated into Version 1: "...越来越大的继承财产的权力" (...the increasing right to inherit wealth) or correctly translated into Version 2: "...继承所得财富带来的越来越大的权力" (...the increasing power brought about by the wealth that is inherited). Both of these versions make semantic sense. However, if subjects employ syntactic processing, the sentence is likely to be correctly translated, because "inherited" here is an adjective rather than a verb, used to modify the noun "wealth". In contrast, "The inherited wealth is about 100 million dollars" (SEM AMB $_-$ SYN AMB $_-$) is likely to be accurately translated only as "继承所得的财产大约是100百万美元" (The wealth that is inherited amounts to approximately 100 million dollars), as only this version makes semantic sense.

In Pair 6 (Question 11 and Question 12), participants were tasked with making a true or false judgment based on a pair of sentences. To ensure comparability, both sentences shared the same syntactic structure: NP + NP. "The price managers" in Question 11 and "the movie Jashon" in Question 12 serve as the main stimuli. If participants primarily rely on semantic processing rather than syntactic processing, they are likely to either mistakenly believe there are "price managers" or correctly understand that there are none, based on the sentence in Question 11. Both interpretations are semantically plausible. Conversely, participants are unlikely to erroneously assume there is "movie Jashon" based on the sentence in Question 12, because the term "movie Jashon" sounds semantically peculiar.

Questions 1, 3, 5, 7, 9, and 11 are identified as semantically ambiguous but syntactically unambiguous, categorized as Group 1 in the subsequent analysis. Questions 2, 4, 6, 8, 10, and 12 are identified as both semantically and syntactically unambiguous, categorized as Group 2.

*2.5 Procedure*

To ensure the validity and reliability of the data obtained, this experiment avoided recruiting participants through instruction from tutors, as this approach often yields reluctant and haphazard responses. Instead, participants were recruited through various channels, including WeChat Moments (a social media platform), Tree Hole (a university forum), and word of mouth. Out of concern that some participants might join the experiment solely for the incentive and could potentially introduce noise into the data by providing random responses, the decision was made to conduct the experiment face-to-face, one-on-one, through interviews rather than utilizing an automated program or questionnaire. While this approach might have introduced some undesired effects, such as stress, on the subjects, it was strongly believed that the benefits far outweighed the drawbacks.

Given that the experiment was conducted during the COVID-19 outbreak, interviews were conducted using computers, tablets, or smartphones from home. Participants were allowed to use their own devices at a location and time of their choosing, but they were not permitted to use any tools to aid in solving the problems presented in the experiment. While the hardware used varied among participants, the software remained consistent. All participants, including the researcher, utilized Tencent Meeting, Version 3.24.2(410), for visual and vocal communication. The researcher provided a brief introduction to the experiment and then presented the experimental sentences to each subject using PowerPoint. The researcher recorded the subjects' responses before moving on to the next question.

## 2.6 Statistical analysis

The data were analyzed utilizing SPSS 20.0.0, GraphPad Prism 8.0.2 (263) and Cursor 0.41.3. Given the meticulous collection process involving face-to-face, one-on-one interviews, we deemed preprocessing unnecessary prior to statistical analysis. Notably, no missing values were identified among the 140 valid samples. Next, only a few outliers were present, notably within Group 1 from the perspective of correctness (see Figure 4a), where the median is 4, the first quartile is 3, the third quartile is 5, the maximum is 6, and the minimum is 0. However, the outliers were not attributed to data collection errors or measurement inaccuracies. Instead, we interpreted them as instances of natural variability and genuine situations. Consequently, we opted to retain these data for statistical analysis. All data concerning outlier, normality, descriptive analysis, and inferential analysis of the 140 samples are provided in the Raw Data File and the Processed Data File, accessible at https://figshare.com/articles/dataset/Raw_Data_File_and_Processed_Data_File/26065912.

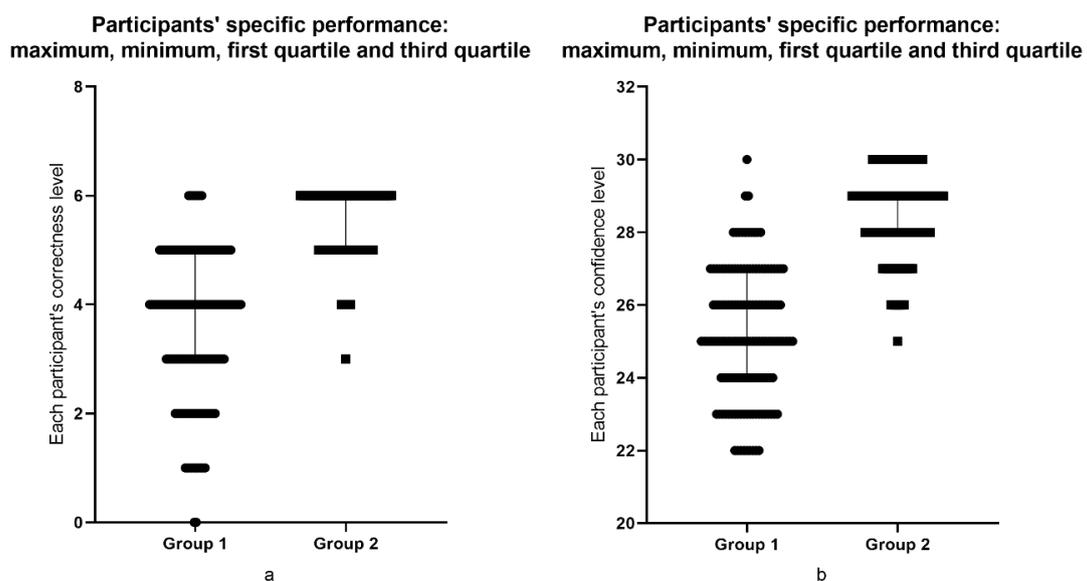

Figure 4 a. The 140 participants' performance in terms of correctness, b. The 140 participants' performance in terms of confidence

Statistical analysis begins with descriptive statistics. For confirming the hypothesis, this study compared the performance of participants on Group 1 and that on Group 2 from two perspectives: correctness and confidence. This aims to determine whether they exhibited better performance in Group 2 compared to Group 1 based on the means.

Subsequently, inferential statistical analysis was employed to ascertain whether a significant difference existed in performance between these two groups. There are many methods for testing significance, including the t-test, which is suitable for comparing the means of two groups, especially when the sample size is small and the data is approximately normally distributed (Ravid 2020). ANOVA is appropriate for comparing the mean differences among multiple groups (Johnson and Wichern 2018). The chi-square test is commonly used to examine the relationship between categorical variables or to test whether the data conforms to an expected theoretical distribution (Starnes and Tabor 2018). These statistical methods fall under parametric statistics, which require the data to meet conditions such as maintaining a normal distribution or having no outliers. However, as per the Anderson-Darling test of normality (Berlinger 2021), the data of the present experiment pertaining to correctness and confidence across both groups did not adhere to a normal distribution (see Table 1). Considering this non-normality primarily stemmed from the limited number of questions within each group rather than by sample selection bias, we opted against normalizing this data. Instead, we chose to employ the statistical method appropriate for non-normally distributed data: the Wilcoxon signed ranks test (Ramsey and Schafer 2013), which does not require the assumption of normality.

| Group | A2 | P-value | Passed normality test? |
|---|---|---|---|
| Correctness level of Group 1 | 4.064 | <0.05 | No |
| Correctness level of Group 2 | 17.83 | <0.05 | No |
| Confidence level of Group 1 | 2.283 | <0.05 | No |
| Confidence level of Group 2 | 5.522 | <0.05 | No |

Table 1 Anderson-Darling test of normality on Group 1 and Group 2 in terms of correctness and confidence

Finally, 95% confidence intervals for the effect size are provided to illustrate the magnitude of the differences. Even if a result is statistically significant (with a p-value less than 0.05), its effect size might be very small, meaning the practical impact may not be noteworthy. Effect size helps to avoid over-reliance on statistical significance (p-value) and instead focuses on the actual practical importance of the findings. There are many methods for calculating effect size. Pearson's r is used to measure the correlation between two continuous variables (Field 2024). η² and Partial η² are used in analysis of variance (ANOVA) to assess the proportion of variance explained by the independent variable (McBride and Cutting 2021). Cohen's d is commonly used to measure the effect size between the means of two groups (Agresti and Finlay 2018). However, these methods are not the best suited for non-parametric tests. Therefore, this study employs the Fisher z transformation to calculate the 95% confidence interval for the effect size of this experiment (Gravetter et al. 2017). Software like SPSS and GraphPad Prism cannot directly calculate the 95% confidence interval for effect size, so we wrote a corresponding Python script (see Figure 5b) using Cursor 0.41.3 based on the five steps of the calculation method found in the literature (see Figure 5a). The standards for effect size corresponding to the r values are as follows: a small effect is indicated by $|r|\approx 0.1$, a medium effect by $|r|\approx 0.3$, and a large effect by $|r| \approx 0.5$ and above. The python application, named Effect_Size.py, is accessible at https://figshare.com/articles/software/Effect_Size_py/27215403?file=49758849.

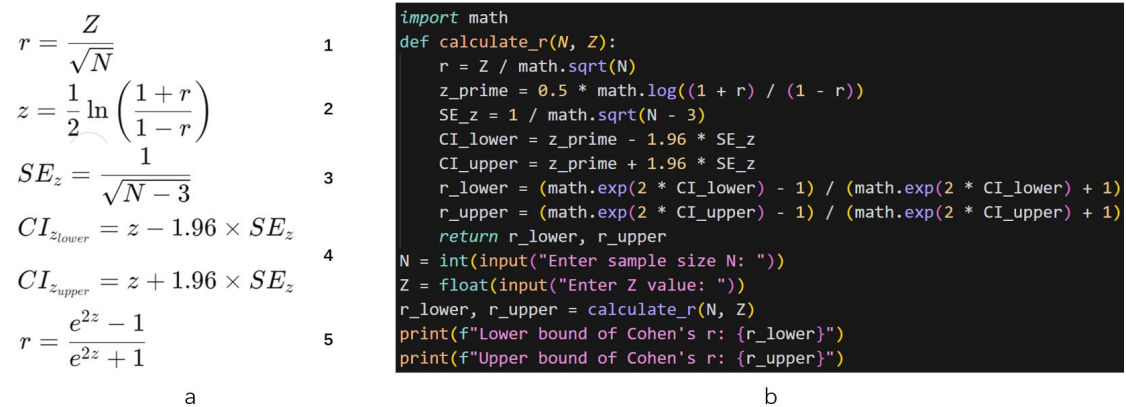

**Figure 5 a.** Calculation method using Fisher z transformation for the 95% confidence interval of the effect size, b. Corresponding Python script

If it was determined not only that subjects in Group 2 performed better than those in Group 1 based on descriptive statistical analysis, but also that there was a statistically significant difference in p-value and a medium to large effect size in inferential statistical analysis, the hypothesis, which posits that Chinese learners of English underutilize syntactic processing for comprehending English sentences, would be confirmed.

## 3 Results

### 3.1 Descriptive statistical results

The descriptive statistical results for Group 1 (SEM AMB $_+$ SYN AMB $_-$) and Group 2 (SEM AMB $_-$ SYN AMB $_-$) in terms of correctness and confidence respectively, are as follows (see Figure 6).

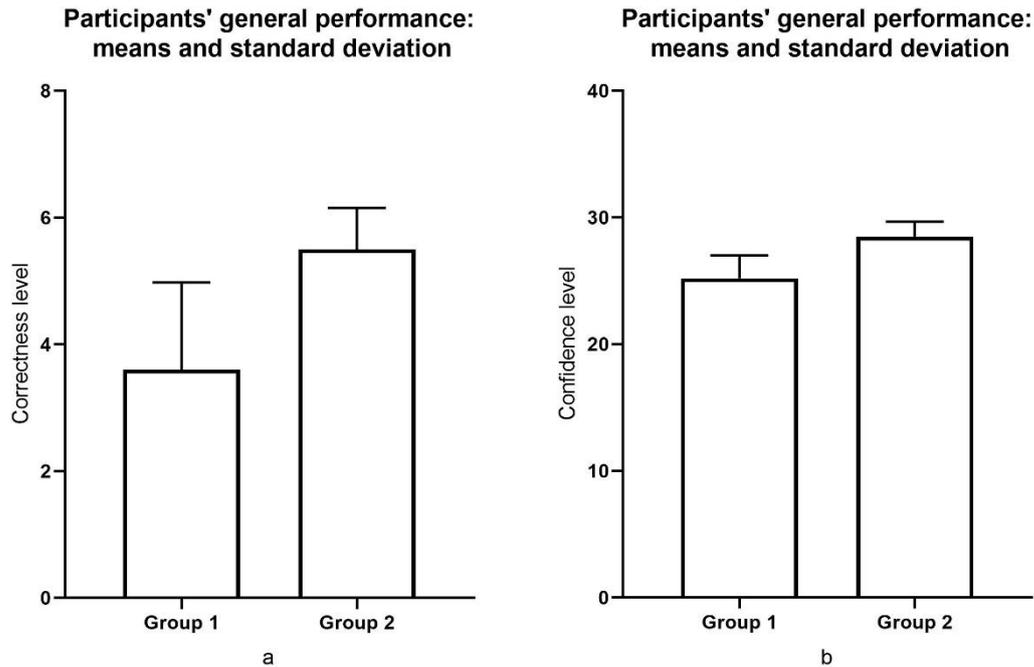

Figure 6 a. The 140 participants' general performance including means and standard deviation on Group 1 (SEM AMB + SYN AMB -) and Group 2 (SEM AMB - SYN AMB -) from the perspective of correctness; b. The 140 participants' general performance including means and standard deviation on Group 1 (SEM AMB + SYN AMB -) and Group 2 (SEM AMB - SYN AMB -) from the perspective of confidence

The Y-axis of Figure 6a indicates the number of questions correctly answered by the subjects in each group, ranging from 0 to 6. The mean correctness level for Group 1 (SEM AMB + SYN AMB -) among the 140 subjects was 3.6, accompanied by a standard deviation of 1.377. In contrast, for Group 2 (SEM AMB - SYN AMB -), the mean correctness level was 5.5, with a standard deviation of 0.652. The Y-axis of Figure 6b indicates the total confidence level with which the subjects responded for all the questions of each group, ranging from 0 to 30. The mean confidence level for Group 1 (SEM AMB + SYN AMB -) was 25.21, with a standard deviation of 1.793. For Group 2 (SEM AMB - SYN AMB -), the mean confidence level was 28.5, with a standard deviation of 1.154. Figure 6a and Figure 6b collectively illustrate a consistent trend where subjects demonstrated superior performance on Group 2 (SEM AMB - SYN AMB -) compared to Group 1 (SEM AMB + SYN AMB -).

### 3.2 Inferential statistical results

According to the two-tailed Wilcoxon signed ranks test, examining correctness levels for Group 1 (SEM AMB + SYN AMB -) versus Group 2 (SEM AMB - SYN AMB -), the results show *positive ranks*=126, *negative ranks*=0, and *ties*=14 ($p<0.001$) (see Figure 7). This indicates that approximately 90% (126 out of 140 subjects) performed worse when sentence meaning was ambiguous. Conversely, none of the subjects (0 out of 140) showed improvement, while 10% (14 out of 140 subjects) performed at the same level. The observed difference in performance between Group 1 (SEM AMB + SYN AMB -) and Group 2 (SEM AMB - SYN AMB -) from the perspective of correctness was significant, with a *p-value* (2-tailed)=0.000 (*p-value*<0.001). Furthermore, this significant difference had a large effect size, with a 95% confidence interval ranging from -0.877 to -0.773 and an absolute value greater than 0.5.

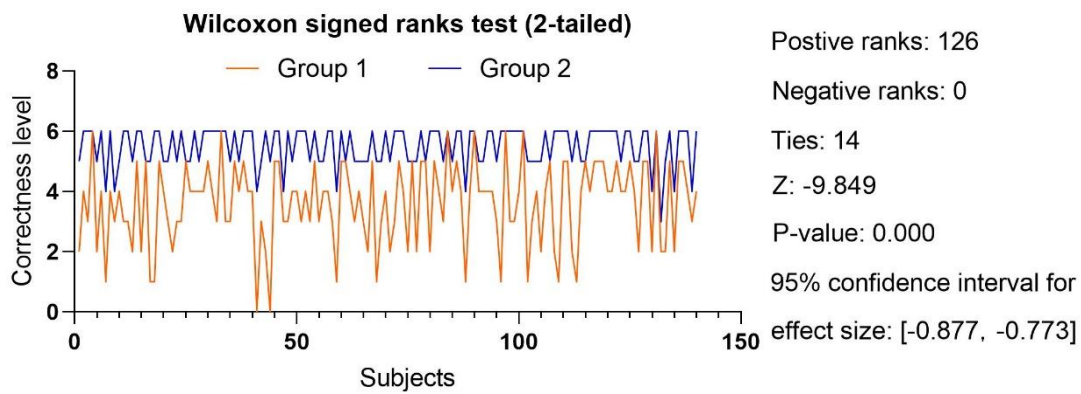

Figure 7 The Wilcoxon signed ranks test (2-tailed) from the perspective of correctness for Group 1 (SEM AMB + SYN AMB -) versus Group 2 (SEM AMB - SYN AMB -)

According to the two-tailed Wilcoxon signed ranks test, examining confidence levels for Group 1 (SEM AMB + SYN AMB -) versus Group 2 (SEM AMB - SYN AMB -), the results reveal *positive ranks*=135, *negative ranks*=1, and *ties*=4 (see Figure 8). This signifies that approximately 96.4% of the total subjects (135 out of 140 individuals) experienced a decline in performance when sentence meaning was ambiguous. There was only one exception (around 0.7%), while 4 subjects (about 2.9%) showed no significant change in their performance. The observed difference in performance between Group 1 (SEM AMB + SYN AMB -) and Group 2 (SEM AMB - SYN AMB -) from the perspective of confidence was also significant, with a *p-value* (2-tailed)=0.000 (*p-value*<0.001). Furthermore, this significant difference had a large effect size, with a 95% confidence interval ranging from -0.897 to -0.809 and an absolute value greater than 0.5.

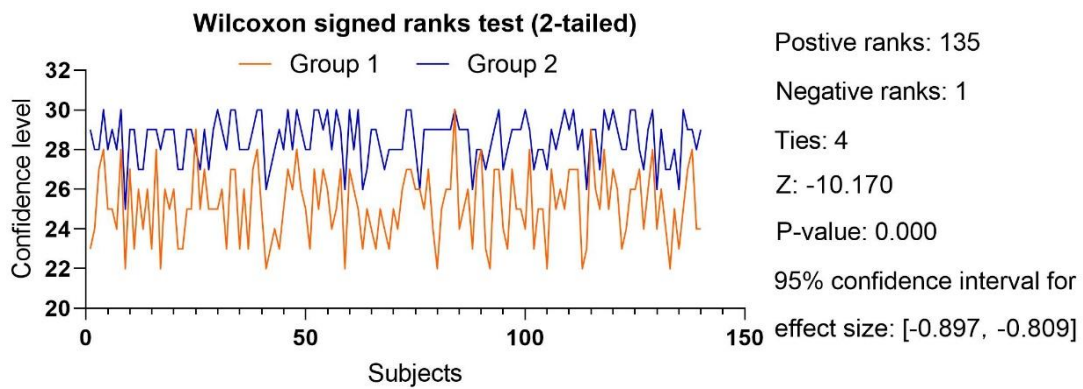

Figure 8 The Wilcoxon signed ranks test (2-tailed) from the perspective of confidence for Group 1 (SEM AMB + SYN AMB -) versus Group 2 (SEM AMB - SYN AMB -)

Based on the results of both descriptive and inferential statistical analyses, the hypothesis that Chinese learners of English underutilize syntactic processing for comprehending English sentences is valid.

# 4    Discussion

The experimental results indicate that participants showed superior performance in answering questions in Group 2 compared to those in Group 1, both in terms of accuracy and confidence, with a higher mean in both measures. Furthermore, this difference is statistically significant, with a large effect size observed in both cases. These findings suggest that Chinese learners of English underutilize syntactic processing for comprehending English sentences.

Theoretically, the conclusion reconfirms and advances Author's findings that syntactic processing often follows sentence comprehension among Chinese learners of English because the underutilization of parsing in sentence comprehension is a reasonable outcome of an ineffective parsing method, such as one that relies on prior sentence comprehension.

Practically, the conclusion highlights a deficiency in the English learning process of Chinese learners. By neglecting L2 syntactic features, learners may impede their ability to develop optimal strategies for processing L2 input (Smith 1993). To address this deficiency, this present study further explores the ways Chinese learners of English underutilize parsing in English sentence understanding and the reasons it is underutilized. This exploration aims to help Chinese learners of English integrate parsing into their comprehension process, thereby enhancing their overall understanding.

## *4.1 Ways Chinese learners of English underutilize syntactic processing in English sentence comprehension*

### *4.1.1   Syntactic-clue-based parsing and trial-and-error-based parsing*

The cue-based retrieval theory suggests that we exploit syntactic clues in the specialized task of sentence parsing (Lewis and Vasishth 2005). For example, we may employ the syntactic clue "that" to determine that "know" is the main verb within the sentence "You know that I am right". In this study, parsing based on obvious syntactic clues is denoted as "syntactic-clue-based parsing" (SCBP).

However, in many cases, there are no obvious syntactic cues, and thereby, ambiguities may arise. For instance, take the sentence "Those forced to exercise their smiling muscles reacted enthusiastically to funny cartoons" as an example. In this case, it becomes challenging to determine which verb, "forced" or "reacted", functions as the main verb compared to the previous case, if we do not determine it based on comprehending the sentence beforehand. Syntactically ambiguous constructions are prevalent across human languages (Hsieh et al. 2009). The researchers employed the stop-making-sense paradigm and eye-tracking during reading to investigate the processing of a syntactic construction. Their findings provided evidence for a limited parallel processor for addressing ambiguities that maintains multiple syntactic analyses across several words of a sentence when the structures are each supported by the available constraints. This approach involving the strategy of trial and error in syntactic processing aligns with several constraint-based models (MacDonald et al. 1994; Spivey and Tanenhaus 1998). These models posit that syntactic representations are initially activated in parallel, and then a single analysis is swiftly selected based on support from various constraints. In the case of "Those forced to exercise their smiling muscles reacted enthusiastically to funny cartoons", we may have two parallel trials: 1) "forced" as the main verb, and 2) "reacted" as the main verb. We then use available constraints to figure out the error, concluding that "forced" is not the main verb because there is no object after this transitive verb. Trial and error is a significant strategy ubiquitously applied to solve almost any problem in our life (Eysenck and Keane 2004; Klahr and Simon 2001). In the context of this study, parsing through trial and error is termed "trial-and-error-based parsing" (TEBP).

*4.1.2 TEBP semantically and TEBP syntactically*

Trial and error is utilized in parsing both semantically and syntactically, particularly in sentences with ambiguity. In the semantic context, trial and error involves testing different combinations of word meanings, assessing their semantic coherence through common sense, and determining their comprehensibility. Parsing results are then derived from successful trials that yield a coherent semantic interpretation based on semantic component definition, such as the main verb within a sentence referring to the verb that expresses the primary action or state of being within a sentence (see Figure 9a).

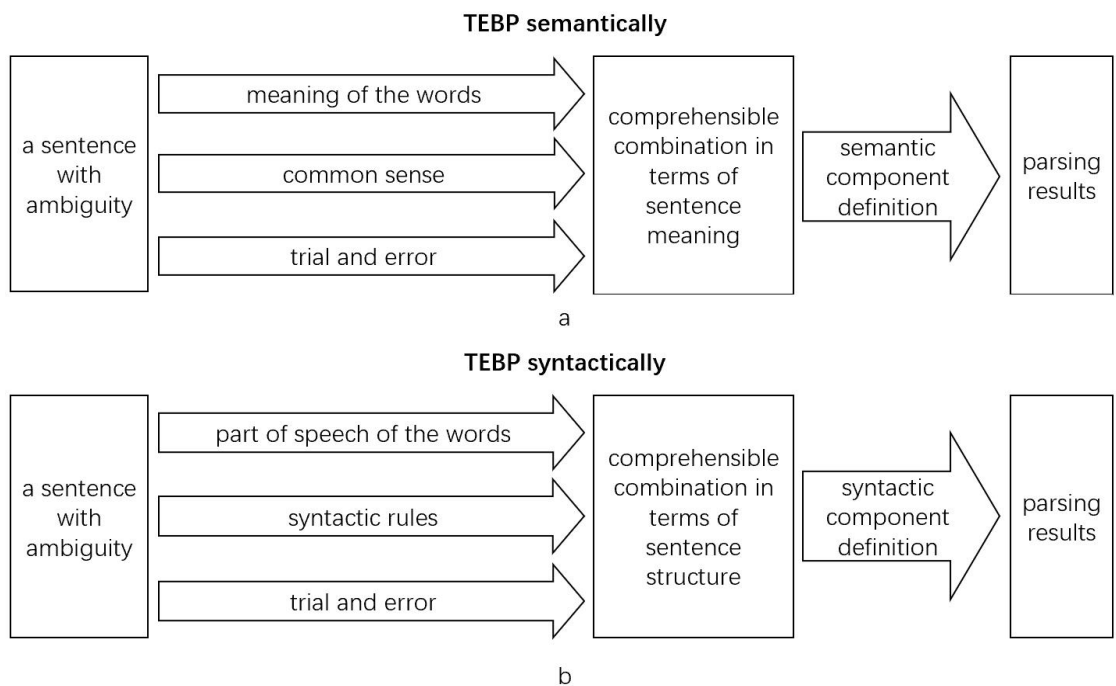

Figure 9 a. TEBP semantically, b. TEBP syntactically

Similarly, trial and error can be employed in parsing through a syntactic approach. In this approach, one can explore various combinations of the parts of speech of words, evaluating their syntactic viability according to syntactic rules to ascertain their comprehensibility. Consequently, parsing results are derived according to syntactic component definition, such as the main verb referring to the verb in the main clause (see Figure 9b).

*4.1.3   Partial underutilization and complete underutilization of syntactic processing*

Unlike TEBP syntactically and SCBP, TEBP semantically diverges by not incorporating syntactic rules (see Figure 9a). Specifically, TEBP semantically involves determining sentence components like subject and object. However, this is accomplished without reliance on syntactic rules. (**In this current study, syntactic processing/parsing includes both applying syntactic rules and identifying sentence components.**) Therefore, TEBP semantically is viewed as a representation of partial underutilization of syntactic processing.

On the other hand, some individuals forgo all forms of parsing, including SCBP, TEBP semantically, TEBP syntactically, etc., and instead comprehend sentences by simply combining lexical meanings with a trial-and-error strategy. This approach bypasses both the application of syntactic rules and the identification of sentence components. Such phenomenon is called complete underutilization of syntactic processing in this study.

**4.2 Reasons Chinese learners of English underutilize syntactic processing in English sentence comprehension**

*4.2.1   Reasons for partial underutilization* of syntactic processing

As discussed in Section 4.1.1 and Section 4.1.2, when obvious syntactic cues are absent, Chinese learners of English usually resort to TEBP approaches, semantically or syntactically. However, the disparity in the effort required to master TEBP semantically versus TEBP syntactically is substantial, as illustrated in Figure 9. TEBP semantically operates on common sense, incurring no additional effort to grasp. Conversely, TEBP syntactically relies on syntactic rules, including sentence patterns, demanding significant effort for proficiency. Nevertheless, in many situations, both TEBP approaches—semantically and syntactically—prove effective for comprehending sentences. Therefore, it is rational for many Chinese learners of English to favor TEBP semantically, the typical representation of partial underutilization of syntactic processing, over its syntactic counterpart. This explains partial underutilization of syntactic processing in English sentence comprehension by Chinese learners of English.

*4.2.2 Reasons for complete underutilization of syntactic processing*

Based on the discussion in Section 4.1.1 and Section 4.1.2, there are presently at least three parsing methods: SCBP, TEBP semantically, and TEBP syntactically, without considering the hybrid of them. Some of these parsing methods necessitate sentence comprehension before syntactic processing and its results, such as TEBP semantically and TEBP syntactically, while others permit sentence comprehension following syntactic processing and its results, such as SCBP (see Figure 10).

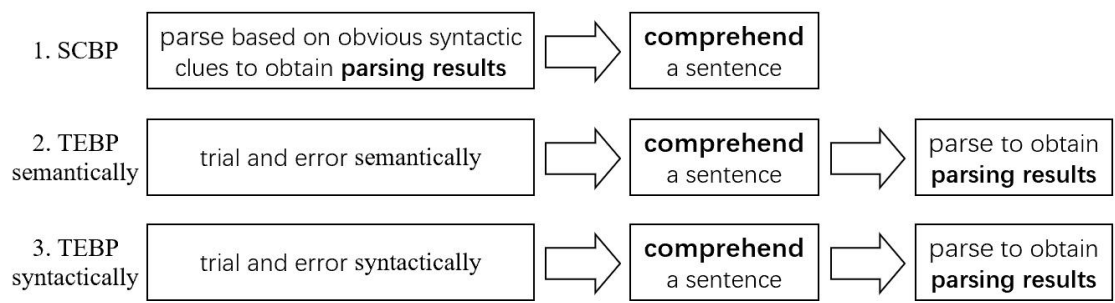

Figure 10 Three parsing methods illustrating different time course of syntactic processing with its results and sentence comprehension

Considering how TEBP semantically and TEBP syntactically function, whenever trial and error is employed, Chinese learners of English parse a sentence to obtain the parsing results after comprehending it, either semantically or syntactically. It is rational for them to halt parsing at the point of sentence comprehension. This strategic pause is utilized to conserve cognitive resources during real-time reading comprehension scenarios. Ideally, the complete underutilization occurs only when SCBP is infeasible, that is, when there are no obvious syntactic clues. However, it might evolve into unconditional complete underutilization spontaneously, as humans tend to make black-and-white decisions. This view is supported by empirical evidence from the performance of Chinese learners of English in identifying the main verb when there are obvious syntactic clues (Author 2024). This explains complete underutilization of syntactic processing in English sentence comprehension by Chinese learners of English.

## 5 Conclusion

This current study demonstrates that Chinese learners of English underutilize syntactic processing when comprehending English sentences, either partially or completely. A detailed exploration of their parsing strategies reveals that a trial-and-error approach results in a complete underutilization of syntactic processing. And the advantage of employing TEBP semantically over TEBP syntactically elucidates the partial underutilization of syntactic parsing. The revelation of the reasons behind the weak role of syntax utilization in English sentence understanding by Chinese learners of English establishes a robust foundation for developing a novel parsing method that fully integrates syntactic processing into sentence comprehension, thereby enhancing cognitive synergy.

One potential solution is to minimize the reliance on trial and error in syntactic processing. Trial and error, although deeply ingrained in our thinking and part of our intuitive knowledge (Klahr and Simon 1999), is an ineffective method. Researchers in many fields (Chaikittisilp and Okubo 2021; Nemeroff 2020), including language education (Ranta 2022), aim to replace it with more effective alternatives. Future research will focus on developing a specific parsing method for English that operates independently of trial and error. Such an approach could enable parsing to precede sentence comprehension, thereby avoiding underutilization by Chinese learners of English.

There are some limitations to this research. The conclusions presented here are currently limited to Chinese learners of English who utilize parsing for English sentence comprehension. It remains unclear to what extent these findings can be generalized to other second-language learning contexts, such as Japanese learners of French using parsing for French sentence comprehension, or even to first-language processing, such as American individuals applying syntactic processing for understanding English sentences. This uncertainty makes it difficult to determine whether the patterns of parsing underutilization are specific to Chinese English learners or reflect a broader trend across various language learning scenarios. Future research could explore whether the results from Chinese English learners apply to other language learning contexts, with the methodologies and achievements of this research serving as a paradigm.